\documentclass[10pt,twocolumn,letterpaper]{article}

\usepackage[pagenumbers]{iccv} 

\usepackage{multirow}
\usepackage{color}
\usepackage{colortbl}
\usepackage{xcolor}
\usepackage{fontawesome5}
\usepackage{xspace}

\newcommand{\icoyes}{\textcolor{forestgreen}{\faIcon{check-circle}}\xspace}

\newcommand{\icono}{\textcolor{ashgrey}{\faIcon{times-circle}}\xspace}
\definecolor{iccvblue}{rgb}{0.21,0.49,0.74}
\usepackage[pagebackref,breaklinks,colorlinks,allcolors=iccvblue]{hyperref}

\definecolor{mygray}{rgb}{.80,.80,.80}
\definecolor{myblue}{rgb}{0.94, 0.95, 1.0}
\definecolor{mylightblue}{rgb}{0.96, 0.97, 1.0}
\definecolor{forestgreen}{rgb}{0.0, 0.5, 0.0}
\definecolor{ashgrey}{rgb}{0.7, 0.75, 0.71}
\definecolor{darkorange}{rgb}{1.0, 0.55, 0.0}

\usepackage{tcolorbox}

\definecolor{myboxcolor}{RGB}{245,245,245} 
\definecolor{myframe}{RGB}{0,0,128} 

\newtcolorbox{mybanner}{
  colback=myboxcolor,
  colframe=myframe,
  boxrule=1pt, 
  left=1pt,
  right=1pt,
  top=1pt,
  bottom=1pt,
}

\newtcolorbox{mybody}{
  colback=myboxcolor,
  colframe=myframe,
  boxrule=1pt, 
  left=1pt,
  right=1pt,
  top=1pt,
  bottom=1pt,
}
\def\paperID{} 
\def\confName{ICCV}
\def\confYear{2025}

\title{GeoGround: A Unified  Large Vision-Language Model \\ for Remote Sensing Visual Grounding}

\author{Yue Zhou\textsuperscript{1}
\and
Mengcheng Lan\textsuperscript{1}
\and
Xiang Li\textsuperscript{2} 
\and 
Litong Feng\textsuperscript{5}
\and
Yiping Ke\textsuperscript{1} 
\and
\qquad\qquad
Xue Jiang\textsuperscript{3}
\and
Qingyun Li\textsuperscript{4}
\and
Xue Yang\textsuperscript{3}
\and
Wayne Zhang\textsuperscript{5}
\and
\small{
\textsuperscript{1}Nanyang Technological University,  \textsuperscript{2}University of Reading} \\
\small{\textsuperscript{3}Shanghai Jiao Tong University, \textsuperscript{4}Harbin Institute of Technology}\\
\small{\textsuperscript{5}SenseTime Research}}

\begin{document}
\maketitle
\begin{abstract}
Remote sensing (RS) visual grounding aims to use natural language expression to locate specific objects (in the form of the bounding box or segmentation mask) in RS images, enhancing human interaction with intelligent RS interpretation systems. Early research in this area was primarily based on horizontal bounding boxes (HBBs), but as more diverse RS datasets have become available, tasks involving oriented bounding boxes (OBBs) and segmentation masks have emerged. In practical applications, different targets require different grounding types: HBB can localize an object's position, OBB provides its orientation, and mask depicts its shape. However, existing specialized methods are typically tailored to a single type of RS visual grounding task and are hard to generalize across tasks. In contrast, large vision-language models (VLMs) exhibit powerful multi-task learning capabilities but struggle to handle dense prediction tasks like segmentation. This paper proposes GeoGround, a novel framework that unifies support for HBB, OBB, and mask RS visual grounding tasks, allowing flexible output selection. Rather than customizing the architecture of VLM, our work aims to elegantly support pixel-level visual grounding output through the Text-Mask technique. We define prompt-assisted and geometry-guided learning to enhance consistency across different signals.  Experimental results show that GeoGround demonstrates strong performance across four RS visual grounding tasks, matching the performance of specialized methods on multiple benchmarks. Code available at \href{https://github.com/zytx121/GeoGround}{https://github.com/zytx121/GeoGround}.
\end{abstract}    
\section{Introduction}
\label{sec:intro}

In the remote sensing (RS) community, the early visual grounding task~\cite{zhan2023rsvg,sun2022visual} specifically refers to the location of specific objects in terms of horizontal bounding boxes (HBBs), given a satellite image and related text query. With increasing availability of the RS dataset~\cite{xia2018dota,li2020object,sun2022fair1m}, researchers have started to use oriented bounding boxes (OBBs)~\cite{kuckreja2024geochat} or segmentation masks~\cite{yuan2024rrsis} to more accurately depict the referred objects. RS visual grounding enables humans to interact with computers in a more intuitive manner, which has enormous promise for improving the efficiency of intelligent RS interpretation systems~\cite{wang2024trustworthy}.

\begin{figure}[!t]
	\begin{center}             
        \includegraphics[width=1\linewidth]{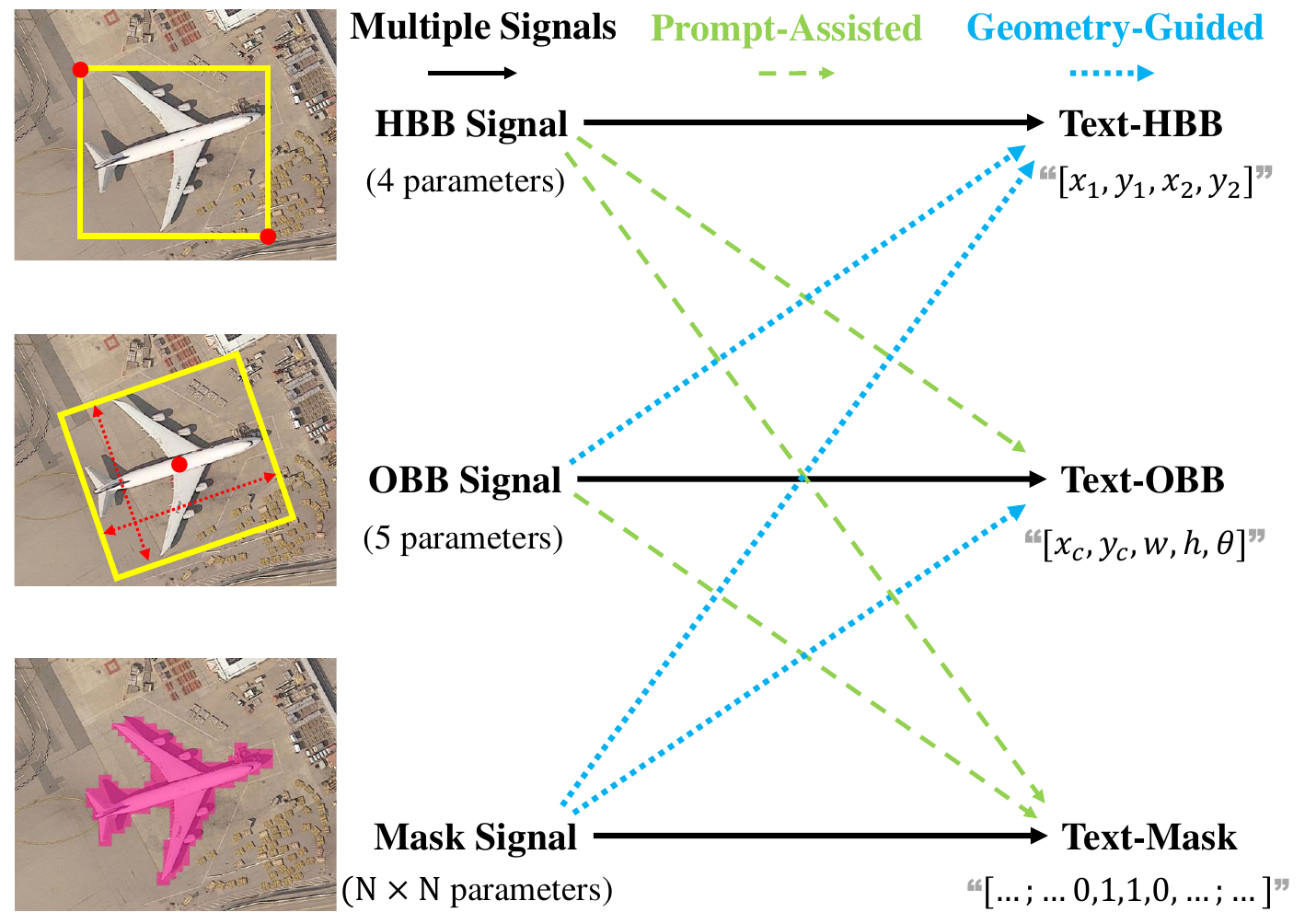}
    \end{center}
    \vspace{-0.5cm}
    \caption{An overview of the hybrid supervision of GeoGround.}
    \label{fig:hybrid}   
\vspace{-0.6cm}
\end{figure}

Most existing RS visual grounding~\cite{zhan2023rsvg,sun2022visual} and referring segmentation~\cite{yuan2024rrsis,liu2024rotated} methods are designed with task-specific modules and loss functions. Models based on HBB typically adopt loss functions in object detection tasks, such as Smooth L1, while models based on mask often use loss functions from semantic segmentation tasks, such as pixel-level cross-entropy. Enabling multi-task learning~\cite{volpi2018deep} in such models not only requires modifications to the network but also necessitates careful tuning of the weightings between various loss functions, making the process quite challenging. Although large vision-language models (VLMs)~\cite{liu2023llava,bai2023qwen,chen2023minigpt,chen2023internvl} can support multiple multimodal RS tasks simultaneously by using a unified text regression loss function, they struggle with pixel-level tasks such as segmentation. This is because, as the output module of a VLM, the large language model (LLM) can only generate textual data and cannot produce output in the image modality~\cite{chen2021pix2seq}.

To address these challenges, we propose GeoGround, an elegant VLM that seamlessly unifies visual grounding tasks at the HBB, OBB, and pixel-level RS. Our key innovation lies in converting box-level and pixel-level signals into textual sequences, enabling the model to train diverse visual grounding tasks within a unified training pipeline. Specifically, we propose the Text-Mask paradigm, which distills and compresses the information embedded in the mask into a compact text sequence that can be efficiently learned by VLMs. Additionally, we introduce hybrid supervision, as shown in Fig. \ref{fig:hybrid}, which incorporates prompt-assisted learning (PAL) and geometry-guided learning (GGL) to fine-tune the model using three types of signals, ensuring output consistency and enhancing the model’s understanding of the relationships between different grounding types.

To support GeoGround training and promote the development of visual RS grounding, we introduce refGeo, a large-scale RS visual grounding instruction-following dataset. refGeo consolidates four existing visual grounding datasets from RS~\cite{sun2022visual,zhan2023rsvg,kuckreja2024geochat,li2024vrsbench} and introduces a new aerial vehicle visual grounding dataset.

In summary, our key contributions are as follows:

\begin{itemize}
  \item We propose GeoGround, a novel VLM framework that unifies box-level and pixel-level RS visual grounding tasks while maintaining its inherent dialogue and image understanding capabilities.
  \item We introduce refGeo, the largest RS visual grounding instruction-following dataset, consisting of 161k image-text pairs and 80k RS images.
  
  \item We conduct extensive experiments on various RS visual grounding tasks, providing valuable insights for future RS VLM research and opening new avenues for research in RS visual grounding.

\end{itemize}

\section{Related Work}
\label{sec:relatedwork}

\begin{figure*}
	\begin{center}             
        \includegraphics[width=1\linewidth]{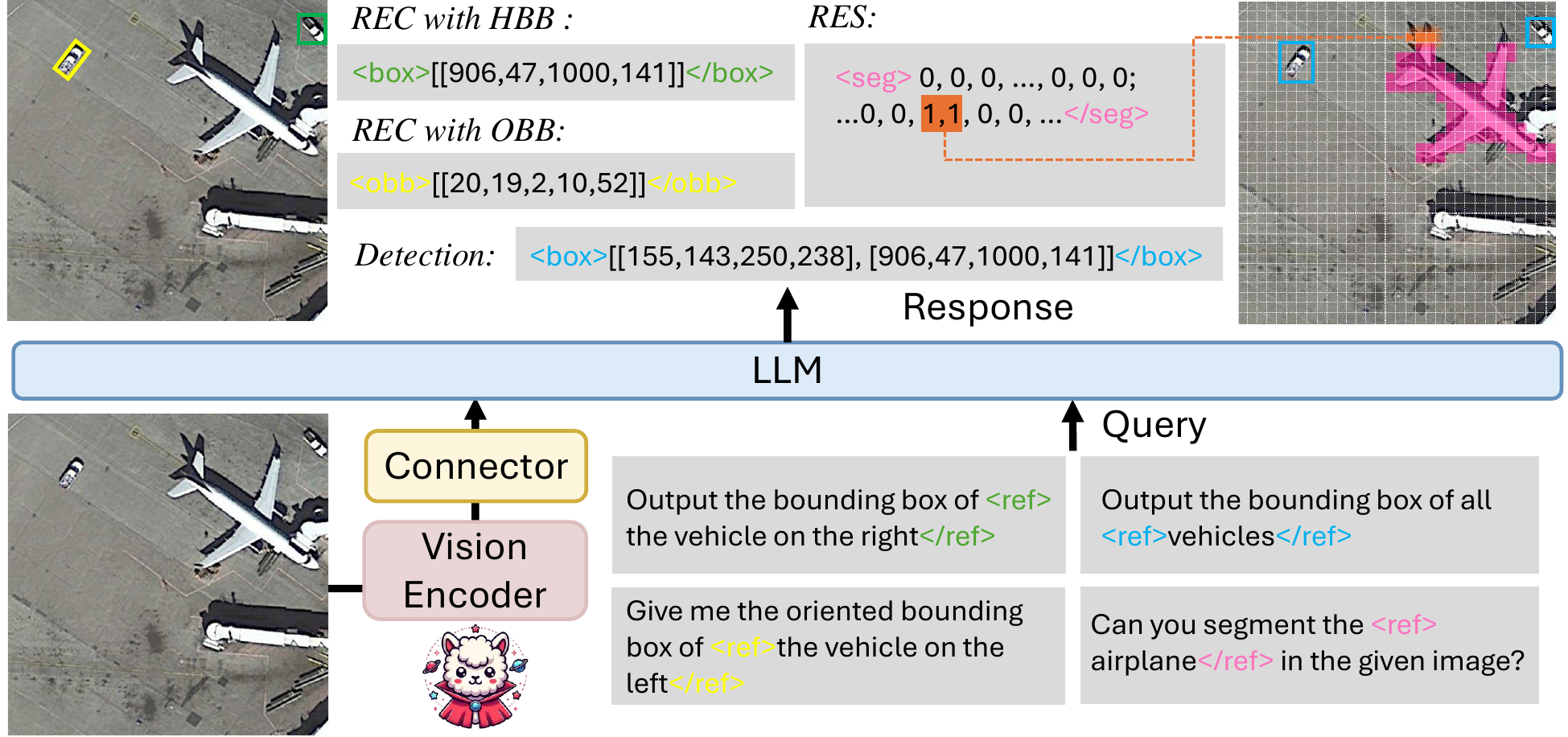}
    \end{center}
    \vspace{-0.5cm}
    \caption{An overview of GeoGround -- the first model to unify box-level and pixel-level visual grounding tasks in remote sensing.}
    \label{fig:geoground}   
\end{figure*}

\paragraph{Remote Sensing Referring Detection and Segmentation.}

Compared to multimodal tasks like image captioning~\cite{zhang2021global,li2021recurrent}, text-image retrieval~\cite{mikriukov2022deep}, and visual question answering (VQA)~\cite{lobry2020rsvqa} in RS, research on referring detection is a novel task with limited research. It was first introduced by GeoVG~\cite{yuan2024rrsis}, which proposed the first RS visual grounding dataset. MGVLF~\cite{zhan2023rsvg} leverages multiscale visual features and multi-granularity textual embeddings to address the scale variation in RS images. LQVG~\cite{lan2024lqvg} propose a language query-based Transformer framework for RSVG. LPVA~\cite{li2024lgpa} achieves precise attention on referred objects by adjusting visual features with progressive attention. RS referring segmentation is also in its early stages due to the challenges mentioned earlier. It was first introduced by RefSegRS~\cite{yuan2024rrsis}, which proposed a new dataset and baseline model. Recently, the transformer-based method RMSIN~\cite{liu2024rotated} proposed an adaptive rotated convolution to tackle the issues arising from the scale variation and orientations prevalent in aerial imagery. FIANet~\cite{lei2024rrsis} proposed a fine-grained image-text alignment module to better discriminative multimodal representation. However, these two types of models have always been studied separately, hindering progress in the field. In this paper, we unify these two tasks within a single framework, allowing them to share data and architecture.

\paragraph{Generalist VLMs.}

Several efforts have been made to equip VLMs with visual grounding capabilities in the domain of natural images~\cite{chen2021pix2seq,chen2023shikra,wang2024visionllm,peng2023kosmos,lai2024lisa,zhang2023next,wu2024visionllm}. For instance, Shikra~\cite{chen2023shikra} directly textualizes HBB to support visual grounding tasks, but its discrete coordinate output is inadequate for pixel-level tasks. LISA~\cite{lai2024lisa} addresses this by incorporating a mask decoder to handle the RES task, while NExT-Chat~\cite{zhang2023next} expands this paradigm by adding two decoders to support both box and mask outputs. In contrast, our approach elegantly unifies box-level and pixel-level visual grounding tasks based on general-purpose VLM, eliminating the need for additional encoders or decoders.

\paragraph{Remote Sensing VLMs} have shown promising results in image-level tasks such as scene classification, image captioning, and VQA~\cite{wang2024skyscript,zhang2024rs5m,li2024vrsbench}. However, works~\cite{kuckreja2024geochat,muhtar2024lhrs,pang2024h2rsvlm,zhang2024earthgpt} on object-level tasks such as RS visual grounding remains comparatively unexplored. GeoChat~\cite{kuckreja2024geochat} constructs a new dataset using OBB annotation and proposes the first RS visual grounding model based on OBB. However, the unsuitable choice of OBB angle representation limits its performance. Moreover, the small scale of the RS visual grounding dataset selected by LHRS-Bot~\cite{muhtar2024lhrs} and H$^2$RSVLM~\cite{pang2024h2rsvlm} hampers their generalization ability on this task. To address these issues, we introduce refGeo, a large-scale RS visual grounding dataset with multi-type annotations. For each type of annotation, we perform a systematic exploration to identify the most suitable format.

\section{GeoGround}

The architecture of GeoGround is highly streamlined, consisting of only a visual encoder (CLIP-ViT~\cite{radford2021learning}), a connector (two-layer MLPs), and an LLM (Vicuna 1.5~\cite{zheng2023judging}), without introducing additional encoders or decoders. \cref{fig:geoground} illustrates the framework of our proposed model, which can flexibly output HBBs, OBBs, or segmentation masks based on user instructions. In addition to single-object outputs, the model is also capable of handling multi-object outputs.

\subsection{Signal Textualization}

To train three types of visual grounding tasks with a unified data pipeline, we textualize the three grounding supervision signals into three corresponding text strings. We refer to this process as signal textualization, which serves as the cornerstone of our method.

\paragraph{Text-HBB and Text-OBB} are generated by directly converting numerical coordinates into text sequences~\cite{chen2023shikra}. Specifically, the coordinates are normalized, multiplied by the resolution, and then rounded. The resulting numbers are separated by commas and enclosed in parentheses, as illustrated in \cref{fig:geoground}. In GeoGround, we set the resolution of the Text-HBB to 1000, allowing for more precise localization of small objects in RS images. Compared to Text-HBB, Text-OBB includes an additional angle parameter. Since there are various angle representations of OBB, the meaning of the first four numbers differs. Based on experiments, we adopt the long side 90-degree representation~\cite{zhou2022mmrotate} in GeoGround, where the angle ranges from $0$ to $90$ degrees. To ensure that these values align with the angle in terms of range, we set the resolution of Text-OBB to 100 by default.

\paragraph{Text-Mask} should be generated by converting the mask into text sequences. However, this conversion is challenging due to the inherent differences between the image and text modalities. Inspired by Text4Seg~\cite{anonymous2024textseg}, we propose a novel Text-Mask paradigm that treats the segmentation mask as text. Specifically, we downsample the mask into an N$\times$N grid, where the object region is labeled 1 and the background region 0, as shown in \cref{fig:geoground}. This results in a binary matrix that approximately represents the object's location and shape. Higher resolution improves shape precision but results in longer text sequences, increasing training difficulty and slowing inference. To further reduce the token length required to represent a mask, we employ R-RLE~\cite{anonymous2024textseg} to compress redundant text sequences. It significantly reduces the length of Text-Mask and accelerates inference speed, without compromising performance. For RS
visual grounding datasets, a resolution of 32 enables Text-Mask to effectively represent most objects.

\subsection{Hybrid Supervision}

We propose a hybrid supervision that simultaneously utilizes Text-HBB, Text-OBB, and Text-Mask to comprehensively enhance the visual grounding capabilities of VLMs. First, we adopt a basic supervised learning paradigm to train three types of visual grounding tasks, as follows:

\begin{equation}
\begin{aligned}
\boldsymbol{t} = \mathcal{F}(\mathbf{I}, \boldsymbol{q})
\end{aligned}
\label{Equ:pal}
\end{equation}
where $\mathcal{F}$ denotes the LLM of our model, $\mathbf{I}$ represents the image embedding, 
$\boldsymbol{q}$ represents the query text embedding. 
$\boldsymbol{t}$ can represent the Text-HBB, Text-OBB, and Text-Mask, respectively. Next, we define the following two auxiliary tasks to establish connections between the different signals.

\paragraph{Prompt-Assisted Learning} refers to completing visual grounding with the help of additional prompts, such as predicting the OBB of an object based on its known HBB. This process can be understood as an increase in information entropy and aims to help the model acquire the ability to generate dense signals from sparse ones. Since dense signals contain more information than sparse signals, this process still requires the model to extract additional information from the image to bridge the gap between the signals. PAL can be expressed by the following equation:
\begin{equation}
\begin{aligned}
\boldsymbol{t}_{dense}=\mathcal{F}(\mathbf{I}, \{\boldsymbol{q}, \boldsymbol{t}_{sparse}\})
\end{aligned}
\label{Equ:pal}
\end{equation}
where $\boldsymbol{t}_{sparse}$ represents the sparse textualized signal, which can be either Text-HBB or Text-OBB here. $\boldsymbol{t}_{dense}$ represents the textualized signal that is denser than $\boldsymbol{t}_{sparse}$.

\begin{figure}[!t]
	\begin{center}             
        \includegraphics[width=1\linewidth]{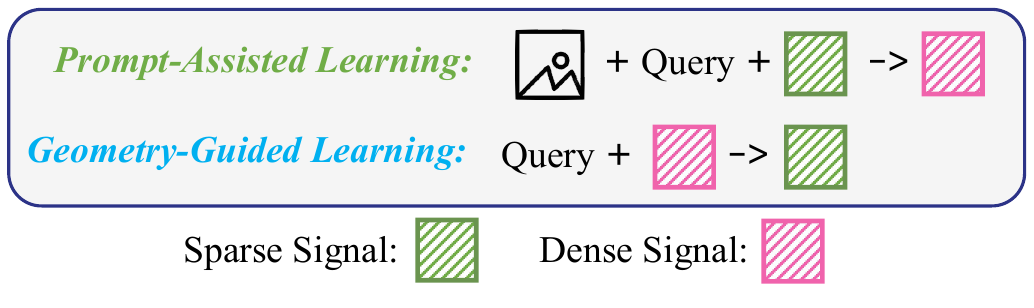}
    \end{center}
    \vspace{-0.4cm}
    \caption{Illustrations of PAL and GGL. The concepts of sparse and dense signals are relative. For example, the sparse signal corresponds to HBB, while the dense signal corresponds to OBB.}
    \label{fig:pal_ggl}          
\vspace{-0.4cm}
\end{figure}

\paragraph{Geometry-Guided Learning} converts dense signals into sparse ones guided by geometric knowledge, reducing information entropy. This means that GGL does not require the image as input; the transformation process can be achieved solely based on geometric knowledge. For example, the HBB that encloses an OBB can be obtained by calculating its four corner points' maximum and minimum values. GGL can be expressed as:
\begin{equation}
\begin{aligned}
\boldsymbol{t}_{sparse}=\mathcal{F}(\{\boldsymbol{q}, \boldsymbol{t}_{dense}\})
\end{aligned}
\label{Equ:gm}
\end{equation}
where $\boldsymbol{t}_{dense}$ denotes the dense textualized signal, which can be either Text-OBB or Text-Mask. Fig. \ref{fig:pal_ggl} presents a demo of PAL and GGL. Similarly to existing VLMs, GeoGround is supervised solely by the text regression loss.

\paragraph{BBox Consistency Score.}

Ideally, the model outputs for the same object should have a similar enclosing bounding box. However, the positions of the HBB, OBB, and mask outputs may differ. To assess prediction consistency, we propose the BBox Consistency Score (BCS):
\begin{equation}
\begin{aligned}
BCS = \frac{1}{3} \Big( &\, \text{IoU}(\boldsymbol{s}_{hbb}, f_{obb2hbb}(\boldsymbol{s}_{obb})) \\
                        &+ \text{IoU}(\boldsymbol{s}_{hbb}, f_{mask2hbb}(\boldsymbol{s}_{mask})) \\
                        &+ \text{IoU}(f_{obb2hbb}(\boldsymbol{s}_{obb}), f_{mask2hbb}(\boldsymbol{s}_{mask})) \Big)
\end{aligned}
\label{Equ:bcs}
\end{equation}
where $\boldsymbol{s}_{hbb}$, $\boldsymbol{s}_{obb}$, and $\boldsymbol{s}_{mask}$ represent HBB, OBB, and mask signals, respectively. IoU denotes the Intersection over Union. $f_{obb2hbb}$ and $f_{mask2hbb}$ represent the functions that compute the enclosing HBB from the OBB and mask, respectively. The BCS ranges from 0 to 1. When the model predictions are completely consistent, the BCS equals 1.

\subsection{Dataset}
\label{sec:dataset_creation}

To address the issue of limited generalization capability in VLMs caused by the relatively small size of existing RS visual grounding datasets, we present a large-scale RS referring expression comprehension dataset, refGeo. It integrates most existing RS visual grounding datasets. The details of each dataset are provided in \cref{tab:total_dataset}. Since both GeoChat~\cite{kuckreja2024geochat} and VRSBench~\cite{li2024vrsbench} use DIOR~\cite{li2020object} image data, which overlap with DIOR-RSVG~\cite{zhan2023rsvg}, we remove the samples corresponding to images that appear in the DIOR-RSVG test and validation sets from the GeoChat and VRSBench training set to prevent data leakage.  Moreover, we propose a new aerial vehicle visual grounding dataset using unmanned aerial vehicles.

\begin{table}
\resizebox{\columnwidth}{!}{
  \centering
  \setlength{\tabcolsep}{2pt}
  \begin{tabular}{lcccc}
    \toprule
    
    Dataset & Format   & \# Refers    & GSD (m) & Image Width  \\
    \midrule
    RSVG~\cite{sun2022visual} & HBB & 5,505  & 0.24$\sim$4.88 & 1,024    \\
    DIOR-RSVG~\cite{zhan2023rsvg} & HBB & 27,133  & 0.5$\sim$30 & 800   \\
    GeoChat~\cite{kuckreja2024geochat} & OBB & 63,883  & 0.3$\sim$0.8 & 600$\sim$1,024    \\
    VRSBench~\cite{li2024vrsbench} & OBB & 38,689  & 0.1$\sim$30 & 512  \\
    Aerial vehicle dataset (Ours) & OBB & 26,465  & 0.007$\sim$0.04 & 4,000 \\
    \rowcolor{myblue} \textbf{refGeo }  & \textbf{Mixed} & \textbf{161,675} & \textbf{0.007$\sim$30} & \textbf{512$\sim$4,000}  \\
    \bottomrule
  \end{tabular}}
  \vspace{-0.2cm}
    \caption{List of datasets used to create our referring expression instruction set for GeoGround VLM training. To ensure diversity, we include RS visual grounding datasets with varying ground sampling distance (GSD) and sizes.}
    \label{tab:total_dataset}
\end{table}


\begin{table*}[!t]
\setlength{\tabcolsep}{3pt}
\resizebox{\textwidth}{!}{
\begin{tabular}{lcccccccc}
\toprule
 Model    & DIOR-RSVG-Test & DIOR-RSVG-Val & RSVG-Test & RSVG-Val & GeoChat* & VRSBench* & AVVG & AVG \\ \midrule
\multicolumn{9}{l}{\hfill \textit{Specialized Models} } \\
\midrule
\textcolor{gray}{MGVLF} \cite{zhan2023rsvg} & \textcolor{gray}{76.78}   & -    &   & -    & -   & -  & -  & -     \\
\textcolor{gray}{GeoVG} \cite{sun2022visual} & -   & -    & \textcolor{gray}{59.40}  & \textcolor{gray}{58.20}    & -   & -  & -  & -     \\
\midrule
\multicolumn{9}{l}{\hfill \textit{Generalist VLMs} } \\
\midrule
InternVL2-8B \cite{chen2023internvl} & 14.42  & 12.99  & 0.16 & 0.67  & 9.91   & 5.47  & 0.34  & 6.28    \\
InternVL2-40B \cite{chen2023internvl} & 15.06  & 14.87  & 0.41  & 0.67   & 21.13  & 13.64  & 8.23  & 10.57     \\
Qwen-VL \cite{bai2023qwen} & 32.22  & 32.01   & 2.04  & 4.66   & 35.36   & 31.07  & 0.31  & 19.66   \\
Qwen2-VL \cite{wang2024qwen2} & 44.25  & 43.32 & 20.13  & 19.15   & 30.92   & 32.88  & 17.73 & 29.77     \\
\midrule
\multicolumn{9}{l}{\hfill \textit{Remote Sensing VLMs} } \\
\midrule
GeoChat \cite{kuckreja2024geochat} & 24.05   & 23.35   & 2.04  & 3.08   & 22.74   & 11.52 & 0.28 & 12.44     \\
LHRS-Bot \cite{muhtar2024lhrs} & 17.59   & 17.04  & 1.56  & 0.95  & 3.25  & 1.19  & 0.00  & 5.94     \\
H$^2$RSVLM  \cite{pang2024h2rsvlm} & 48.04   & -   & -  & -    & -   & -  & -  & -     \\
EarthGPT \cite{zhang2024earthgpt} & 76.65   & -   & -  & -    & -   & -  & -  & -     \\
\midrule
\multicolumn{9}{l}{\hfill \textit{Supervised Fine-Tuning on refGeo} } \\
\midrule
Qwen-VL (sft) & 58.76   & 58.65    & 10.59  & 12.99  & 41.75  & 47.38  & 9.53  & 34.24    \\
GeoChat (sft) & 61.96   & 60.27   & 14.67  & 16.32   & 56.99  & 51.36 & 11.52  & 39.01     \\
LLaVA-1.5-7B (sft) & 65.98   & 64.46    & 20.95 & 19.98  & 63.76   & 57.17  & 15.05  & 43.91     \\
\rowcolor{mylightblue}GeoGround (N=32) & 76.42  & 76.18  & 26.57  & 26.73  & 68.65  & 65.35  & 21.34  & 51.61    \\
\rowcolor{myblue}GeoGround (N=16) & \textbf{77.73}  & \textbf{77.18}  & \textbf{26.65}  & \textbf{27.64}  & \textbf{70.24}  & \textbf{66.04}  & \textbf{21.58}  & \textbf{52.44}    \\
\rowcolor{myblue} Improvement over Runner-up $\uparrow$ & \textcolor{red}{1.4\%} & \textcolor{red}{19.7\%} & \textcolor{red}{27.2\%} & \textcolor{red}{38.3\%} & \textcolor{red}{10.2\%} & \textcolor{red}{15.5\%} & \textcolor{red}{21.7\%} & \textcolor{red}{19.4\%} \\
\bottomrule
\end{tabular}}
\vspace{-0.2cm}
\caption{Performance (Acc@0.5\%) comparison on 7 benchmarks. $*$ indicates that the test set has been modified. The last row indicates the performance improvement of our method compared to the runner-up VLM, further demonstrating the superiority of the GeoGround.} \label{tab:rec_overall}
\vspace{-0.4cm}
\end{table*}

\section{Experiments}


Our method is based on LLaVA-1.5-7B~\cite{liu2023llava} with the input image resolution fixed at 336$\times$336. We utilize the AdamW optimizer~\cite{loshchilov2017decoupled}, starting with an initial learning rate of 2e-4, followed by a linear decay schedule after a warm-up phase with a 0.03 ratio. To reduce GPU memory consumption, all models are fine-tuned using LoRA with a rank of 64, in conjunction with ZeRO-2 stage memory optimization. All models are trained on 8 NVIDIA V100 GPUs (32GB) with a global batch size of 128 for 5 epochs. The inference batch size is set to 1 for all experiments. 3 RS object detection datasets~\cite{xia2018dota,li2020object,sun2022fair1m} are used during the fine-tuning of GeoGround to enhance its basic visual perception capabilities.

\subsection{Referring Expression Comprehension (REC)}  

\paragraph{Settings.} 
We follow standard evaluation protocols~\cite{pang2024h2rsvlm,li2024vrsbench} and assess the REC task using Acc@0.5 metric. Except for H2RSVLM~\cite{pang2024h2rsvlm} and EarthGPT~\cite{zhang2024earthgpt}, whose metrics are cited in the original articles due to the lack of open source code, the results for the other VLMs are obtained by inference with the official model weights provided. For the GeoChat~\cite{kuckreja2024geochat}, we convert its output OBBs to HBBs.

\paragraph{Results.} 
\cref{tab:rec_overall} compares the performance of GeoGround with 2 specialized models and 8 mainstream VLMs on 7 REC benchmarks. GeoGround achieves the best performance across all benchmarks, surpassing the specialized model on the DIOR-RSVG test set. VLMs fine-tuned on our refGeo dataset, such as Qwen-VL~\cite{bai2023qwen} and GeoChat~\cite{kuckreja2024geochat}, showed significant improvements on the REC task, validating the effectiveness of the scaling law in the field of RS visual grounding. Benefiting from the wide range of image resolutions and GSD in refGeo, the fine-tuned model showed significant performance improvements on datasets with a high proportion of small objects, such as RSVG and AVVG. GeoGround achieves the best performance when the resolution of Text-Mask is set to 16. This could be due to the increased difficulty of training at higher resolutions. Although low resolution leads to coarse masks, they can be seen as attention mechanisms that aid in localizing the approximate object area.

\subsection{REC with OBB}

\paragraph{Settings.}
Following GeoChat~\cite{kuckreja2024geochat}, we also use Acc@0.5 as metric, with the difference being that rotated IoU~\cite{yang2022detecting} is used instead of normal IoU during the calculation.

\paragraph{Results.} 

\cref{tab:me_5} compares the performance of GeoGround with GeoChat and LLaVA-1.5 on three REC benchmarks that provide OBB annotations. The results demonstrate GeoGround's dominance in RS visual grounding tasks based on OBB, further validating the effectiveness of our hybrid supervision approach. Due to the increased number of parameters to learn, this task is more challenging than standard REC, resulting in lower scores on the OBB task compared to the HBB task, even on the same test set.

\begin{table}[!t]
\vspace{-0.2cm}
\resizebox{\columnwidth}{!}{
  \centering
  \setlength{\tabcolsep}{6pt}
  \begin{tabular}{lcccc}
    \toprule

    Method  &GeoChat$^*$ & VRSBench$^*$ & AVVG & AVG \\
    \midrule
     GeoChat  & 31.88 & 11.54  & 0.00 & 14.47 \\
     LLaVA-1.5-7B (sft)   & 51.47 & 43.71  & 12.56 & 35.91\\
\rowcolor{myblue}GeoGround (N=16) & \textbf{59.72} & \textbf{53.22}  & \textbf{13.93}  & \textbf{42.29} \\
    \bottomrule
  \end{tabular}}
  \vspace{-0.2cm}
    \caption{Performance (Acc@0.5) comparison on 3 REC benchmarks that provide OBB annotations.}
    \label{tab:me_5}
    \vspace{-0.2cm}
\end{table}







\subsection{Referring Expression Segmentation (RES)}

\paragraph{Settings.}
We utilize Acc@0.5 and Mean Intersection-over-Union (mIoU) as evaluation metrics, similar to prior studies~\cite{wu2020phrasecut,yuan2024rrsis}. As GeoGround is currently the only RS VLM that supports the RES task, we compare it with three generalist VLMs that possess native segmentation capabilities on the RES task in the RRSIS-D dataset~\cite{liu2024rotated}.

\paragraph{Results.} 

\cref{tab:ae_2} demonstrates that GeoGround exhibits superior performance in the pixel-level RS vision grounding task.  GeoGround (N=32) achieve better results than directly using HBB to prompt SAM even without relying on SAM. Moreover, we attempt to use HBB and coarse mask to prompt SAM \cite{kirillov2023segment}, which allowed GeoGround to achieve results that match the performance of the best RS referring segmentation model \cite{liu2024rotated}. See appendix for more details.


\begin{table}[!t]
\vspace{-0.2cm}
\resizebox{\columnwidth}{!}{
  \centering
  \setlength{\tabcolsep}{8pt}
  \begin{tabular}{lcccc}
    \toprule
    \multirow{2}{*}{Method} & \multicolumn{2}{c}{Acc@0.5} &  \multicolumn{2}{c}{mIoU}\\
    \cline{2-5}
     & Val & Test  & Val & Test \\
    \midrule
    \textcolor{gray}{RSMIN}~\cite{liu2024rotated} & \textcolor{mygray}{74.66} & \textcolor{mygray}{74.26} & \textcolor{mygray}{65.10}  & \textcolor{mygray}{64.20} \\
    LISA~\cite{lai2024lisa} & 27.07  & 24.51 & 27.84  & 26.78 \\  PixelLM~\cite{ren2024pixellm} & 33.46 & 28.81 & 33.89 & 31.65 \\
    NExT-Chat~\cite{zhang2023next} & 28.97 & 26.37 & 26.98 & 24.98 \\  
    
    Qwen2-VL (HBB) + SAM & 44.97  & 44.38 & 42.40 & 41.30 \\
    LLaVA (sft) + SAM & 60.41  & 59.38 & 55.93 & 54.61 \\
    \rowcolor{mylightblue}GeoGround (N=16) & 42.93  & 40.57 & 42.24 & 41.05 \\
    \rowcolor{myblue}GeoGround (N=32) & 63.25  & 60.97 & 56.36  & 54.92 \\\rowcolor{myblue}GeoGround (N=16) + SAM & 68.69  & \textbf{67.50} & \textbf{61.10} & \textbf{60.50} \\
    \rowcolor{myblue}GeoGround (N=32) + SAM & \textbf{68.88}  & 66.06 & 60.80 & 58.93 \\
    \bottomrule
  \end{tabular}}
  \vspace{-0.2cm}
    \caption{Performance comparison of RES task on the RRSIS-D.}
    \label{tab:ae_2}
\vspace{-0.2cm}
\end{table}

\begin{table}[!t]
\resizebox{\columnwidth}{!}{
  \centering
  \setlength{\tabcolsep}{6pt}
  \begin{tabular}{lcccc}
    \toprule

    \multirow{2}{*}{Options} & 
    \multicolumn{3}{c}{Acc@0.5} & \multirow{2}{*}{BCS} \\
    \cline{2-4}
    & HBB & OBB$\rightarrow$HBB &Mask$\rightarrow$HBB & \\
    \midrule
    LLaVA-1.5-7B & 79.84 & - & - & -  \\
    \midrule
    + Multiple Signals & 87.56 & 80.76 & 61.40 & 0.62  \\
    + PAL & 86.76 & 80.82 & 63.24 & 0.63  \\
    \rowcolor{myblue}+ GGL & \textbf{87.87} & \textbf{81.83} & \textbf{64.95} & \textbf{0.64}  \\
    \bottomrule
  \end{tabular}}
  \vspace{-0.2cm}
    \caption{Ablation Study of Hybrid Supervision on RRSIS-D.}
    \label{tab:me_3_1}
\end{table}

\begin{table}[!t]
\vspace{-0.2cm}
\resizebox{\columnwidth}{!}{
  \centering
  \setlength{\tabcolsep}{2pt}
  \begin{tabular}{ccccc}
    \toprule

    HBB$\rightarrow$OBB & HBB$\leftarrow$OBB & Acc@0.5 (HBB) & Acc@0.5 (OBB) & BCS \\
    \midrule
    \icono & \icono  & 60.87 & 58.68 & 0.67   \\
    PAL & \icono  & 61.37 & 59.12 & 0.68   \\
    \icono & PAL & 60.83 & 58.27 & 0.68   \\
    \icono & GGL & 60.60 & 58.81 & 0.69   \\
    PAL & PAL  & 61.18 & 59.53 & 0.69   \\
    \rowcolor{myblue}PAL & GGL  & \textbf{61.39} & \textbf{59.71} & \textbf{0.70}   \\
    \bottomrule
  \end{tabular}}
  \vspace{-0.2cm}
    \caption{Influence of signal consistency on DIOR-RSVG.}
    \label{tab:me_3__}
    \vspace{-0.1cm}
\end{table}

\begin{table}[!t]
\resizebox{\columnwidth}{!}{
  \centering
  \setlength{\tabcolsep}{4pt}
  \begin{tabular}{ccccccc}
    \toprule
    \multirow{2}{*}{HBB} & \multirow{2}{*}{OBB} & \multirow{2}{*}{Mask} & 
    \multicolumn{2}{c}{Acc@0.5 (HBB)} & \multicolumn{2}{c}{Acc@0.5 (OBB)} \\
    \cline{4-7}
    &  &  & DIOR-RSVG & RSVG & VRSBench & GeoChat \\
    \midrule
    \icoyes & \icono & \icono  & 65.98 & 20.95 & 41.45 & 42.95   \\
    \icono & \icoyes & \icono  & 60.75 & 14.51 & 43.71 & 51.47  \\
    \icoyes & \icoyes & \icono  & 68.49 & 22.66 & 48.01 & 55.64   \\
    \rowcolor{myblue}\icoyes & \icoyes & \icoyes  & \textbf{68.61} & \textbf{23.47} & \textbf{49.66} & \textbf{56.43}   \\
    \bottomrule
  \end{tabular}}
  \vspace{-0.2cm}
    \caption{Influence of multiple supervised signals on GeoGround.}
    \label{tab:me_3_}
\vspace{-0.4cm}
\end{table}

\subsection{Ablation Study}
\label{sec:ablation_study}

\paragraph{Effect of Hybrid Supervision.}

\cref{tab:me_3_1} presents the ablation study of the components in our proposed hybrid supervision method. To compute BCS, we first convert both OBB and mask into HBB before calculating Acc@0.5. The results confirm their effectiveness and further highlight the importance of output consistency in improving performance. \cref{tab:me_3__} illustrates the impact of different learning strategies on the performance of VLM. All models are trained on the training set and evaluated on the DIOR-RSVG test set. The results show that PAL improves performance when predicting dense signals from sparse ones, while GM, which requires no visual input, yields better results when predicting sparse signals from dense ones. \cref{tab:me_3_} further explores the effect of multiple signals on model performance. The results show that direct training with three signals can enhance the visual grounding capability of the VLM on HBB and OBB tasks.

\paragraph{Design Options of Text-Mask} 

To our best knowledge, Text4Seg~\cite{anonymous2024textseg} is the only work to attempt to treat masks as text. However, with longer referring expressions, its semantic descriptors become overly redundant. \cref{tab:ae_7} compares the performance of our proposed Text-Mask with Text4Seg on the DIOR-RSVG test set. The HBB prediction is obtained by calculating the bounding box from the boundary of the segmentation mask. Experiments show that mapping the mask to a binary matrix not only reduces the text length of Text4Seg by 40\% but also improves its
performance by 18\%. Since the objects in RS are relatively small, using the nearest downsampling method leads to the loss of mask information for small objects, resulting in significant performance degradation. While increasing the mask quantization resolution can further improve the segmentation accuracy, longer output text sequences increase the inference time and training difficulty.

\begin{table}[!t]
\resizebox{\columnwidth}{!}{
  \centering
  \setlength{\tabcolsep}{3pt}
  \begin{tabular}{lcccc}
    \toprule
    Method & DownSample & Resolution & Length & Acc@0.5 (HBB) \\
    \midrule
    Text4Seg~\cite{anonymous2024textseg} & NEAREST & 16  & 330 & 17.68   \\
    Text4Seg~\cite{anonymous2024textseg} & MaxPooling & 16  & 397 & 43.09  \\
    \rowcolor{mylightblue}Text-Mask & MaxPooling & 16  & 157 & 50.73  \\
    \rowcolor{myblue}Text-Mask & MaxPooling & 24  & 237 & 56.10  \\
    \rowcolor{myblue}Text-Mask & MaxPooling & 32  & 316 & \textbf{57.50} \\
    \bottomrule
  \end{tabular}}
  \vspace{-0.2cm}
    \caption{Discrepancies between Text-Mask and Text4Seg. Length represents the average length of the model outputs.}
    \label{tab:ae_7}
\end{table}




\begin{figure*}[h]
\vspace{-0.2cm}
	\begin{center}             
        \includegraphics[width=1\linewidth]{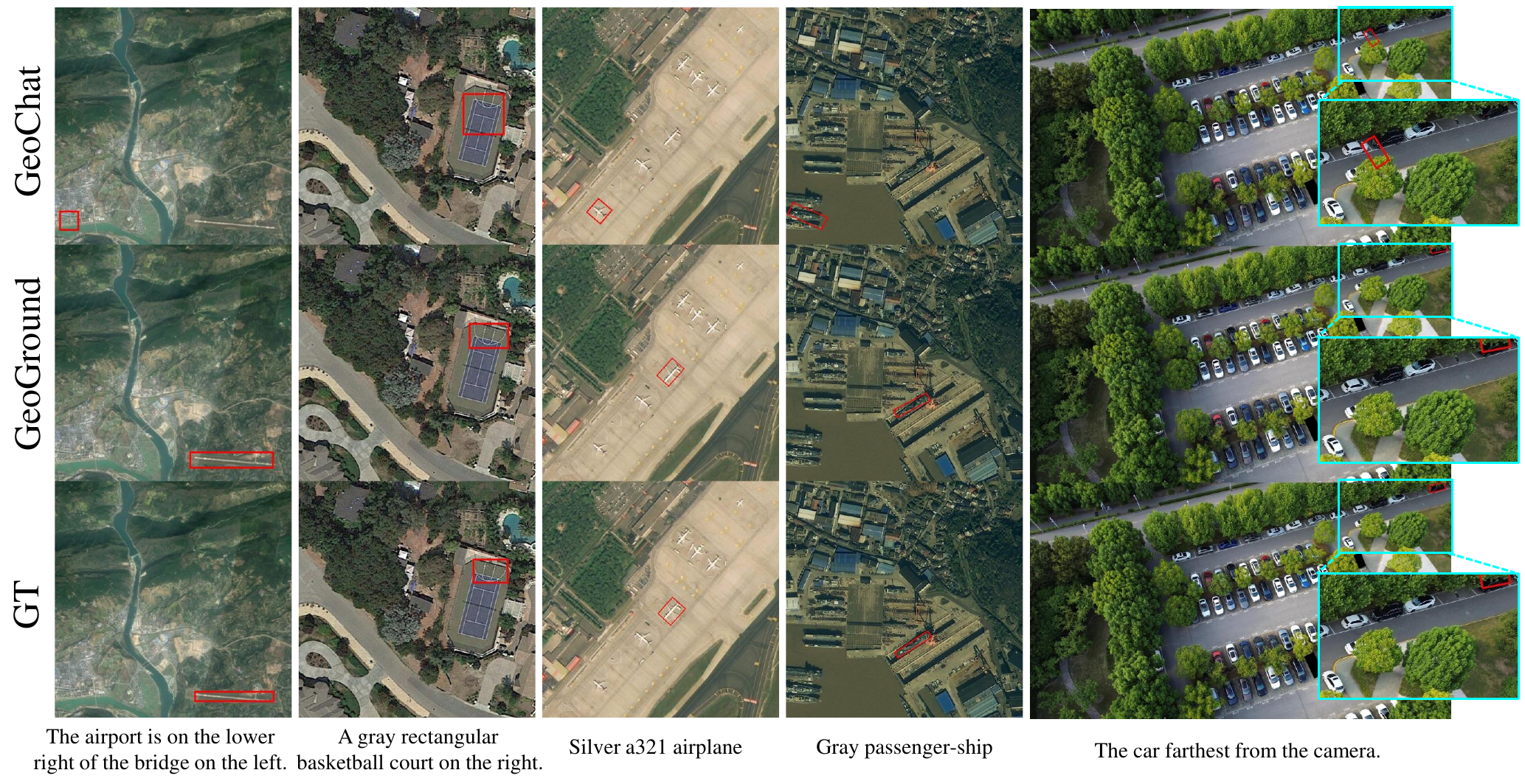}
    \end{center}
\vspace{-0.6cm}
    \caption{Visualizations of GeoGround and GeoChat on the REC task with HBB and OBB.}
    \label{fig:viz_hbb&obb}        
\end{figure*}

\begin{figure*}[h]
\vspace{-0.2cm}
	\begin{center}             
        \includegraphics[width=1\linewidth]{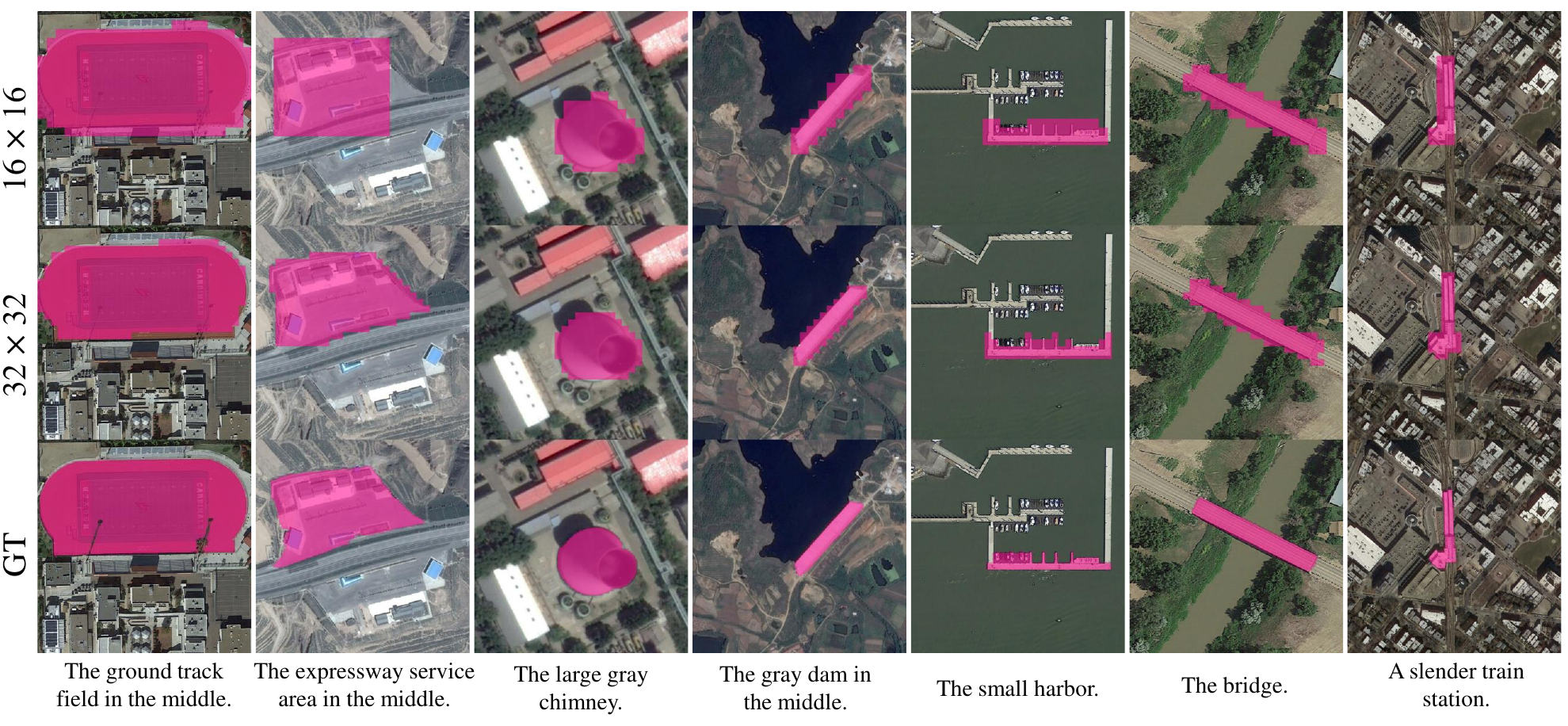}
    \end{center}
\vspace{-0.6cm}
    \caption{Visualizations of GeoGround with different resolutions of Text-Mask on the RRSIS-D test set.}
    \label{fig:viz_seg}   
\vspace{-0.4cm}
\end{figure*}


\subsection{Visualization Examples}

Qualitative comparisons between GeoGround and GeoChat are presented in Fig. \ref{fig:viz_hbb&obb}, which includes HBBs and OBBs. It can be observed that GeoGround consistently demonstrates superior localization accuracy, whether handling simple or relatively complex referential expressions. Additionally, it exhibits 3D spatial understanding, enabling it to infer 3D distances from 2D images. Fig. \ref{fig:viz_seg} compares the performance of GeoGround in the RES task under different resolutions of Text-Mask. When the resolution is 32, although the coarse mask edges still exhibit small jaggedness, the result is already very close to the ground truth. These results fully validate the effectiveness of GeoGround in addressing the pixel-level visual grounding task in RS.

\section{Conclusion}

Although remote sensing (RS) visual grounding tasks using horizontal bounding boxes, oriented bounding boxes, and segmentation masks have progressed, no model has unified these tasks due to framework limitations. To address this, we propose GeoGround, a novel framework that unifies box-level and pixel-level visual grounding tasks in a single model. Instead of adding extra encoders or decoders, GeoGround empowers large vision-language models (VLMs) to perform pixel-level visual grounding by treating the segmentation mask as text using our Text-Mask method. It does not compromise the model's conversational abilities or its image-level understanding capabilities. We also introduce a large-scale RS visual grounding instruction-following dataset, refGeo, that offers a comprehensive benchmark for various visual grounding tasks of RS and serves as a valuable corpus for RS VLMs. Our comprehensive benchmarks and ablation studies provide important insights for the development of VLMs in the RS domain.

{
    \small
    \bibliographystyle{ieeenat_fullname}
    \bibliography{main}

\begin{thebibliography}{44}
\providecommand{\natexlab}[1]{#1}
\providecommand{\url}[1]{\texttt{#1}}
\expandafter\ifx\csname urlstyle\endcsname\relax
  \providecommand{\doi}[1]{doi: #1}\else
  \providecommand{\doi}{doi: \begingroup \urlstyle{rm}\Url}\fi

\bibitem[Bai et~al.(2023)Bai, Bai, Yang, Wang, Tan, Wang, Lin, Zhou, and Zhou]{bai2023qwen}
Jinze Bai, Shuai Bai, Shusheng Yang, Shijie Wang, Sinan Tan, Peng Wang, Junyang Lin, Chang Zhou, and Jingren Zhou.
\newblock Qwen-vl: A frontier large vision-language model with versatile abilities.
\newblock \emph{arXiv preprint arXiv:2308.12966}, 2023.

\bibitem[Chen et~al.(2023{\natexlab{a}})Chen, Zhu, Shen, Li, Liu, Zhang, Krishnamoorthi, Chandra, Xiong, and Elhoseiny]{chen2023minigpt}
Jun Chen, Deyao Zhu, Xiaoqian Shen, Xiang Li, Zechun Liu, Pengchuan Zhang, Raghuraman Krishnamoorthi, Vikas Chandra, Yunyang Xiong, and Mohamed Elhoseiny.
\newblock Minigpt-v2: large language model as a unified interface for vision-language multi-task learning.
\newblock \emph{arXiv preprint arXiv:2310.09478}, 2023{\natexlab{a}}.

\bibitem[Chen et~al.(2023{\natexlab{b}})Chen, Zhang, Zeng, Zhang, Zhu, and Zhao]{chen2023shikra}
Keqin Chen, Zhao Zhang, Weili Zeng, Richong Zhang, Feng Zhu, and Rui Zhao.
\newblock Shikra: Unleashing multimodal llm's referential dialogue magic.
\newblock \emph{arXiv preprint arXiv:2306.15195}, 2023{\natexlab{b}}.

\bibitem[Chen et~al.(2021)Chen, Saxena, Li, Fleet, and Hinton]{chen2021pix2seq}
Ting Chen, Saurabh Saxena, Lala Li, David~J Fleet, and Geoffrey Hinton.
\newblock Pix2seq: A language modeling framework for object detection.
\newblock \emph{arXiv preprint arXiv:2109.10852}, 2021.

\bibitem[Chen et~al.(2023{\natexlab{c}})Chen, Wu, Wang, Su, Chen, Xing, Zhong, Zhang, Zhu, Lu, Li, Luo, Lu, Qiao, and Dai]{chen2023internvl}
Zhe Chen, Jiannan Wu, Wenhai Wang, Weijie Su, Guo Chen, Sen Xing, Muyan Zhong, Qinglong Zhang, Xizhou Zhu, Lewei Lu, Bin Li, Ping Luo, Tong Lu, Yu Qiao, and Jifeng Dai.
\newblock Internvl: Scaling up vision foundation models and aligning for generic visual-linguistic tasks.
\newblock \emph{arXiv preprint arXiv:2312.14238}, 2023{\natexlab{c}}.

\bibitem[Kirillov et~al.(2023)Kirillov, Mintun, Ravi, Mao, Rolland, Gustafson, Xiao, Whitehead, Berg, Lo, et~al.]{kirillov2023segment}
Alexander Kirillov, Eric Mintun, Nikhila Ravi, Hanzi Mao, Chloe Rolland, Laura Gustafson, Tete Xiao, Spencer Whitehead, Alexander~C Berg, Wan-Yen Lo, et~al.
\newblock Segment anything.
\newblock In \emph{Proceedings of the IEEE/CVF International Conference on Computer Vision}, pages 4015--4026, 2023.

\bibitem[Kuckreja et~al.(2024)Kuckreja, Danish, Naseer, Das, Khan, and Khan]{kuckreja2024geochat}
Kartik Kuckreja, Muhammad~Sohail Danish, Muzammal Naseer, Abhijit Das, Salman Khan, and Fahad~Shahbaz Khan.
\newblock Geochat: Grounded large vision-language model for remote sensing.
\newblock In \emph{Proceedings of the IEEE/CVF Conference on Computer Vision and Pattern Recognition}, pages 27831--27840, 2024.

\bibitem[Lai et~al.(2024)Lai, Tian, Chen, Li, Yuan, Liu, and Jia]{lai2024lisa}
Xin Lai, Zhuotao Tian, Yukang Chen, Yanwei Li, Yuhui Yuan, Shu Liu, and Jiaya Jia.
\newblock Lisa: Reasoning segmentation via large language model.
\newblock In \emph{Proceedings of the IEEE/CVF Conference on Computer Vision and Pattern Recognition}, pages 9579--9589, 2024.

\bibitem[Lan et~al.(2024{\natexlab{a}})Lan, Chen, Zhou, Xu, Ke, Wang, Feng, and Zhang]{anonymous2024textseg}
Mengcheng Lan, Chaofeng Chen, Yue Zhou, Jiaxing Xu, Yiping Ke, Xinjiang Wang, Litong Feng, and Wayne Zhang.
\newblock Text4seg: Reimagining image segmentation as text generation.
\newblock \emph{arXiv preprint arXiv:2410.09855}, 2024{\natexlab{a}}.

\bibitem[Lan et~al.(2024{\natexlab{b}})Lan, Rong, Jiao, Gao, and Zhang]{lan2024lqvg}
Meng Lan, Fu Rong, Hongzan Jiao, Zhi Gao, and Lefei Zhang.
\newblock Language query-based transformer with multiscale cross-modal alignment for visual grounding on remote sensing images.
\newblock \emph{IEEE Transactions on Geoscience and Remote Sensing}, 62:\penalty0 1--13, 2024{\natexlab{b}}.

\bibitem[Lei et~al.(2025)Lei, Xiao, Zhang, Li, Shi, and Zhu]{lei2024rrsis}
Sen Lei, Xinyu Xiao, Tianlin Zhang, Heng-Chao Li, Zhenwei Shi, and Qing Zhu.
\newblock Exploring fine-grained image-text alignment for referring remote sensing image segmentation.
\newblock \emph{IEEE Transactions on Geoscience and Remote Sensing}, 63:\penalty0 1--11, 2025.

\bibitem[Li et~al.(2020)Li, Wan, Cheng, Meng, and Han]{li2020object}
Ke Li, Gang Wan, Gong Cheng, Liqiu Meng, and Junwei Han.
\newblock Object detection in optical remote sensing images: A survey and a new benchmark.
\newblock \emph{ISPRS journal of photogrammetry and remote sensing}, 159:\penalty0 296--307, 2020.

\bibitem[Li et~al.(2024{\natexlab{a}})Li, Wang, Xu, Zhong, and Wang]{li2024lgpa}
Ke Li, Di Wang, Haojie Xu, Haodi Zhong, and Cong Wang.
\newblock Language-guided progressive attention for visual grounding in remote sensing images.
\newblock \emph{IEEE Transactions on Geoscience and Remote Sensing}, 62:\penalty0 1--13, 2024{\natexlab{a}}.

\bibitem[Li et~al.(2024{\natexlab{b}})Li, Ding, and Elhoseiny]{li2024vrsbench}
Xiang Li, Jian Ding, and Mohamed Elhoseiny.
\newblock Vrsbench: A versatile vision-language benchmark dataset for remote sensing image understanding.
\newblock \emph{arXiv preprint arXiv:2406.12384}, 2024{\natexlab{b}}.

\bibitem[Li et~al.(2021)Li, Zhang, Gu, Li, Wang, Tang, and Jiao]{li2021recurrent}
Yunpeng Li, Xiangrong Zhang, Jing Gu, Chen Li, Xin Wang, Xu Tang, and Licheng Jiao.
\newblock Recurrent attention and semantic gate for remote sensing image captioning.
\newblock \emph{IEEE Transactions on Geoscience and Remote Sensing}, 60:\penalty0 1--16, 2021.

\bibitem[Liu et~al.(2023)Liu, Li, Wu, and Lee]{liu2023llava}
Haotian Liu, Chunyuan Li, Qingyang Wu, and Yong~Jae Lee.
\newblock Visual instruction tuning.
\newblock \emph{arXiv preprint arXiv:2304.08485}, 2023.

\bibitem[Liu et~al.(2024)Liu, Ma, Zhang, Wang, Ji, Sun, and Ji]{liu2024rotated}
Sihan Liu, Yiwei Ma, Xiaoqing Zhang, Haowei Wang, Jiayi Ji, Xiaoshuai Sun, and Rongrong Ji.
\newblock Rotated multi-scale interaction network for referring remote sensing image segmentation.
\newblock In \emph{Proceedings of the IEEE/CVF Conference on Computer Vision and Pattern Recognition}, pages 26658--26668, 2024.

\bibitem[Lobry et~al.(2020)Lobry, Marcos, Murray, and Tuia]{lobry2020rsvqa}
Sylvain Lobry, Diego Marcos, Jesse Murray, and Devis Tuia.
\newblock Rsvqa: Visual question answering for remote sensing data.
\newblock \emph{IEEE Transactions on Geoscience and Remote Sensing}, 58\penalty0 (12):\penalty0 8555--8566, 2020.

\bibitem[Loshchilov(2017)]{loshchilov2017decoupled}
I Loshchilov.
\newblock Decoupled weight decay regularization.
\newblock \emph{arXiv preprint arXiv:1711.05101}, 2017.

\bibitem[Mikriukov et~al.(2022)Mikriukov, Ravanbakhsh, and Demir]{mikriukov2022deep}
Georgii Mikriukov, Mahdyar Ravanbakhsh, and Beg{\"u}m Demir.
\newblock Deep unsupervised contrastive hashing for large-scale cross-modal text-image retrieval in remote sensing.
\newblock \emph{arXiv preprint arXiv:2201.08125}, 2022.

\bibitem[Muhtar et~al.(2024)Muhtar, Li, Gu, Zhang, and Xiao]{muhtar2024lhrs}
Dilxat Muhtar, Zhenshi Li, Feng Gu, Xueliang Zhang, and Pengfeng Xiao.
\newblock Lhrs-bot: Empowering remote sensing with vgi-enhanced large multimodal language model.
\newblock \emph{arXiv preprint arXiv:2402.02544}, 2024.

\bibitem[Pang et~al.(2024)Pang, Wu, Li, Liu, Sun, Li, Weng, Wang, Feng, Xia, et~al.]{pang2024h2rsvlm}
Chao Pang, Jiang Wu, Jiayu Li, Yi Liu, Jiaxing Sun, Weijia Li, Xingxing Weng, Shuai Wang, Litong Feng, Gui-Song Xia, et~al.
\newblock H2rsvlm: Towards helpful and honest remote sensing large vision language model.
\newblock \emph{arXiv preprint arXiv:2403.20213}, 2024.

\bibitem[Peng et~al.(2023)Peng, Wang, Dong, Hao, Huang, Ma, and Wei]{peng2023kosmos}
Zhiliang Peng, Wenhui Wang, Li Dong, Yaru Hao, Shaohan Huang, Shuming Ma, and Furu Wei.
\newblock Kosmos-2: Grounding multimodal large language models to the world.
\newblock \emph{arXiv preprint arXiv:2306.14824}, 2023.

\bibitem[Radford et~al.(2021)Radford, Kim, Hallacy, Ramesh, Goh, Agarwal, Sastry, Askell, Mishkin, Clark, et~al.]{radford2021learning}
Alec Radford, Jong~Wook Kim, Chris Hallacy, Aditya Ramesh, Gabriel Goh, Sandhini Agarwal, Girish Sastry, Amanda Askell, Pamela Mishkin, Jack Clark, et~al.
\newblock Learning transferable visual models from natural language supervision.
\newblock In \emph{International conference on machine learning}, pages 8748--8763. PMLR, 2021.

\bibitem[Ren et~al.(2024)Ren, Huang, Wei, Zhao, Fu, Feng, and Jin]{ren2024pixellm}
Zhongwei Ren, Zhicheng Huang, Yunchao Wei, Yao Zhao, Dongmei Fu, Jiashi Feng, and Xiaojie Jin.
\newblock Pixellm: Pixel reasoning with large multimodal model.
\newblock In \emph{Proceedings of the IEEE/CVF Conference on Computer Vision and Pattern Recognition}, pages 26374--26383, 2024.

\bibitem[Sun et~al.(2022{\natexlab{a}})Sun, Wang, Yan, Xu, Wang, Diao, Chen, Li, Feng, Xu, et~al.]{sun2022fair1m}
Xian Sun, Peijin Wang, Zhiyuan Yan, Feng Xu, Ruiping Wang, Wenhui Diao, Jin Chen, Jihao Li, Yingchao Feng, Tao Xu, et~al.
\newblock Fair1m: A benchmark dataset for fine-grained object recognition in high-resolution remote sensing imagery.
\newblock \emph{ISPRS Journal of Photogrammetry and Remote Sensing}, 184:\penalty0 116--130, 2022{\natexlab{a}}.

\bibitem[Sun et~al.(2022{\natexlab{b}})Sun, Feng, Li, Ye, Kang, and Huang]{sun2022visual}
Yuxi Sun, Shanshan Feng, Xutao Li, Yunming Ye, Jian Kang, and Xu Huang.
\newblock Visual grounding in remote sensing images.
\newblock In \emph{Proceedings of the 30th ACM International Conference on Multimedia}, pages 404--412, 2022{\natexlab{b}}.

\bibitem[Volpi and Tuia(2018)]{volpi2018deep}
Michele Volpi and Devis Tuia.
\newblock Deep multi-task learning for a geographically-regularized semantic segmentation of aerial images.
\newblock \emph{ISPRS journal of photogrammetry and remote sensing}, 144:\penalty0 48--60, 2018.

\bibitem[Wang et~al.(2024{\natexlab{a}})Wang, Bai, Tan, Wang, Fan, Bai, Chen, Liu, Wang, Ge, et~al.]{wang2024qwen2}
Peng Wang, Shuai Bai, Sinan Tan, Shijie Wang, Zhihao Fan, Jinze Bai, Keqin Chen, Xuejing Liu, Jialin Wang, Wenbin Ge, et~al.
\newblock Qwen2-vl: Enhancing vision-language model's perception of the world at any resolution.
\newblock \emph{arXiv preprint arXiv:2409.12191}, 2024{\natexlab{a}}.

\bibitem[Wang et~al.(2024{\natexlab{b}})Wang, Han, Huang, Zhang, Wang, and Li]{wang2024trustworthy}
Sheng Wang, Wei Han, Xiaohui Huang, Xiaohan Zhang, Lizhe Wang, and Jun Li.
\newblock Trustworthy remote sensing interpretation: Concepts, technologies, and applications.
\newblock \emph{ISPRS Journal of Photogrammetry and Remote Sensing}, 209:\penalty0 150--172, 2024{\natexlab{b}}.

\bibitem[Wang et~al.(2024{\natexlab{c}})Wang, Chen, Chen, Wu, Zhu, Zeng, Luo, Lu, Zhou, Qiao, et~al.]{wang2024visionllm}
Wenhai Wang, Zhe Chen, Xiaokang Chen, Jiannan Wu, Xizhou Zhu, Gang Zeng, Ping Luo, Tong Lu, Jie Zhou, Yu Qiao, et~al.
\newblock Visionllm: Large language model is also an open-ended decoder for vision-centric tasks.
\newblock \emph{Advances in Neural Information Processing Systems}, 36, 2024{\natexlab{c}}.

\bibitem[Wang et~al.(2024{\natexlab{d}})Wang, Prabha, Huang, Wu, and Rajagopal]{wang2024skyscript}
Zhecheng Wang, Rajanie Prabha, Tianyuan Huang, Jiajun Wu, and Ram Rajagopal.
\newblock Skyscript: A large and semantically diverse vision-language dataset for remote sensing.
\newblock In \emph{Proceedings of the AAAI Conference on Artificial Intelligence}, pages 5805--5813, 2024{\natexlab{d}}.

\bibitem[Wu et~al.(2020)Wu, Lin, Cohen, Bui, and Maji]{wu2020phrasecut}
Chenyun Wu, Zhe Lin, Scott Cohen, Trung Bui, and Subhransu Maji.
\newblock Phrasecut: Language-based image segmentation in the wild.
\newblock In \emph{Proceedings of the IEEE/CVF Conference on Computer Vision and Pattern Recognition}, pages 10216--10225, 2020.

\bibitem[Wu et~al.(2024)Wu, Zhong, Xing, Lai, Liu, Wang, Chen, Zhu, Lu, Lu, et~al.]{wu2024visionllm}
Jiannan Wu, Muyan Zhong, Sen Xing, Zeqiang Lai, Zhaoyang Liu, Wenhai Wang, Zhe Chen, Xizhou Zhu, Lewei Lu, Tong Lu, et~al.
\newblock Visionllm v2: An end-to-end generalist multimodal large language model for hundreds of vision-language tasks.
\newblock \emph{arXiv preprint arXiv:2406.08394}, 2024.

\bibitem[Xia et~al.(2018)Xia, Bai, Ding, Zhu, Belongie, Luo, Datcu, Pelillo, and Zhang]{xia2018dota}
Gui-Song Xia, Xiang Bai, Jian Ding, Zhen Zhu, Serge Belongie, Jiebo Luo, Mihai Datcu, Marcello Pelillo, and Liangpei Zhang.
\newblock Dota: A large-scale dataset for object detection in aerial images.
\newblock In \emph{Proceedings of the IEEE conference on computer vision and pattern recognition}, pages 3974--3983, 2018.

\bibitem[Yang et~al.(2022)Yang, Zhang, Yang, Zhou, Wang, Tang, He, and Yan]{yang2022detecting}
Xue Yang, Gefan Zhang, Xiaojiang Yang, Yue Zhou, Wentao Wang, Jin Tang, Tao He, and Junchi Yan.
\newblock Detecting rotated objects as gaussian distributions and its 3-d generalization.
\newblock \emph{IEEE Transactions on Pattern Analysis and Machine Intelligence}, 45\penalty0 (4):\penalty0 4335--4354, 2022.

\bibitem[Yuan et~al.(2024)Yuan, Mou, Hua, and Zhu]{yuan2024rrsis}
Zhenghang Yuan, Lichao Mou, Yuansheng Hua, and Xiao~Xiang Zhu.
\newblock Rrsis: Referring remote sensing image segmentation.
\newblock \emph{IEEE Transactions on Geoscience and Remote Sensing}, 2024.

\bibitem[Zhan et~al.(2023)Zhan, Xiong, and Yuan]{zhan2023rsvg}
Yang Zhan, Zhitong Xiong, and Yuan Yuan.
\newblock Rsvg: Exploring data and models for visual grounding on remote sensing data.
\newblock \emph{IEEE Transactions on Geoscience and Remote Sensing}, 61:\penalty0 1--13, 2023.

\bibitem[Zhang et~al.(2023)Zhang, Zhao, Xie, Zheng, Ji, and Chua]{zhang2023next}
Ao Zhang, Liming Zhao, Chen-Wei Xie, Yun Zheng, Wei Ji, and Tat-Seng Chua.
\newblock Next-chat: An lmm for chat, detection and segmentation.
\newblock \emph{arXiv preprint arXiv:2311.04498}, 2023.

\bibitem[Zhang et~al.(2024{\natexlab{a}})Zhang, Cai, Zhang, Zhuang, and Mao]{zhang2024earthgpt}
Wei Zhang, Miaoxin Cai, Tong Zhang, Yin Zhuang, and Xuerui Mao.
\newblock Earthgpt: A universal multi-modal large language model for multi-sensor image comprehension in remote sensing domain.
\newblock \emph{IEEE Transactions on Geoscience and Remote Sensing}, 2024{\natexlab{a}}.

\bibitem[Zhang et~al.(2021)Zhang, Zhang, Yan, Gao, Fu, and Sun]{zhang2021global}
Zhengyuan Zhang, Wenkai Zhang, Menglong Yan, Xin Gao, Kun Fu, and Xian Sun.
\newblock Global visual feature and linguistic state guided attention for remote sensing image captioning.
\newblock \emph{IEEE Transactions on Geoscience and Remote Sensing}, 60:\penalty0 1--16, 2021.

\bibitem[Zhang et~al.(2024{\natexlab{b}})Zhang, Zhao, Guo, and Yin]{zhang2024rs5m}
Zilun Zhang, Tiancheng Zhao, Yulong Guo, and Jianwei Yin.
\newblock Rs5m and georsclip: A large scale vision-language dataset and a large vision-language model for remote sensing.
\newblock \emph{IEEE Transactions on Geoscience and Remote Sensing}, 2024{\natexlab{b}}.

\bibitem[Zheng et~al.(2023)Zheng, Chiang, Sheng, Zhuang, Wu, Zhuang, Lin, Li, Li, Xing, et~al.]{zheng2023judging}
Lianmin Zheng, Wei-Lin Chiang, Ying Sheng, Siyuan Zhuang, Zhanghao Wu, Yonghao Zhuang, Zi Lin, Zhuohan Li, Dacheng Li, Eric Xing, et~al.
\newblock Judging llm-as-a-judge with mt-bench and chatbot arena.
\newblock \emph{Advances in Neural Information Processing Systems}, 36:\penalty0 46595--46623, 2023.

\bibitem[Zhou et~al.(2022)Zhou, Yang, Zhang, Wang, Liu, Hou, Jiang, Liu, Yan, Lyu, et~al.]{zhou2022mmrotate}
Yue Zhou, Xue Yang, Gefan Zhang, Jiabao Wang, Yanyi Liu, Liping Hou, Xue Jiang, Xingzhao Liu, Junchi Yan, Chengqi Lyu, et~al.
\newblock Mmrotate: A rotated object detection benchmark using pytorch.
\newblock In \emph{Proceedings of the 30th ACM International Conference on Multimedia}, pages 7331--7334, 2022.

\end{thebibliography}
}

\end{document}